\pdfoutput=1

\documentclass{article}

\usepackage[nonatbib, preprint]{neurips_2018}

\usepackage{subfiles}
\usepackage[utf8]{inputenc} 
\usepackage[T1]{fontenc}    
\usepackage{hyperref}       
\usepackage{url}            
\usepackage{booktabs}       
\usepackage{amsfonts}       
\usepackage{nicefrac}       
\usepackage{microtype}      
\usepackage{multirow}
\usepackage{amsmath}

\usepackage{wrapfig}
\usepackage{lipsum}
\usepackage{comment}
\usepackage{subcaption}
\usepackage{graphicx}

\usepackage{xargs}  
\usepackage[pdftex,dvipsnames]{xcolor}
\usepackage[colorinlistoftodos,prependcaption,textsize=scriptsize]{todonotes}
\newcommandx{\note}[2][1=]{\todo[linecolor=red,backgroundcolor=red!25,bordercolor=red,#1]{#2}}

\makeatletter
\newcommand{\printfnsymbol}[1]{%
  \textsuperscript{\@fnsymbol{#1}}%
}

\title{Towards Compact and Robust Deep Neural Networks}

\author{%
  Vikash Sehwag \thanks{Equal contribution.}~~\thanks{Corresponding author. \textit{vvikash@princeton.edu}}\\
  Princeton University \\
  \And
  Shiqi Wang \printfnsymbol{1}\\
  Columbia University \\
  \AND
  Prateek Mittal \\
  Princeton University \\
  \And
  Suman Jana \\
  Columbia University \\
}

\begin{document}

\maketitle
\vspace{-10pt}

\begin{abstract}
Deep neural networks have achieved impressive performance in many applications but their large number of parameters lead to significant computational and storage overheads. Several recent works attempt to mitigate these overheads by designing compact networks using pruning of connections. However, we observe that most of the existing strategies to design compact networks fail to preserve network robustness against adversarial examples. In this work, we rigorously study the extension of network pruning strategies to preserve both benign accuracy and robustness of a network. Starting with a formal definition of the pruning procedure, including pre-training, weights pruning, and fine-tuning, we propose a new pruning method that can create compact networks while preserving both benign accuracy and robustness. Our method is based on two main insights: (1) we ensure that the training objectives of the pre-training and fine-tuning steps match the training objective of the desired robust model (e.g., adversarial robustness/verifiable robustness), and (2) we keep the pruning strategy agnostic to pre-training and fine-tuning objectives. We evaluate our method on four different networks on the CIFAR-10 dataset and measure benign accuracy, empirical robust accuracy, and verifiable robust accuracy. We demonstrate that our pruning method can preserve on average 93\% benign accuracy, 92.5\% empirical robust accuracy, and 85.0\%  verifiable robust accuracy while compressing the tested network by 10$\times$.
\end{abstract}

\label{sec:introduction}

\section{Introduction}
Machine learning, fueled by the recent advances in deep neural networks, has made tremendous progress in many real-world applications~\cite{silver2016masteringGo, esteva2019DLinHealtcare, hinton2012DLSpeechreco, krizhevsky2012imagenet, long2015FCN, ren2015fasterRCNN, wang2017DLProteinFold, plis2014DLMedImaging, bojarski2016Steeringangle, taigman2014deepface}. However, existing deep neural networks face two key challenges. First, existing neural networks are not compact, i.e, they have millions of parameters that lead to high computational and memory storage costs~\cite{canziani2016DLPerformancevsSizeSurvey, han2015nips, li2016l1filterprune}. Such overheads hinder the deployment of these networks in resource-constrained environments such as embedded systems, mobile, or IoT devices~\cite{han2015nips, han2016eie, yao2017deepiot, parashar2017scnn}. The second challenge is the threat of adversarial examples~\cite{goodfellow2014explaining, madry2017towards, szegedy2013intriguing, carlini2017towards, zhang2019tradeoff, athalye2018obfuscated, tencent2019teslaattack}, where an adversary introduce small input perturbations during inference inducing misclassifications. These attacks have sparked a line of research enhancing the security of neural networks against adversarial examples~\cite{tramer2017ensembleadvtrain, madry2017towards, song2018robustdomainadapt, zhang2019tradeoff,shiqi2018efficient,  wang2018mixtrain}. However existing works on improving compactness and adversarial robustness focus on these challenges in isolation. We believe that these two challenges must be considered in conjunction for creating safe, compact, and efficient neural networks. Along this direction, we ask the following key question: \textit{Is it possible to achieve both high compactness and robustness in a single neural network?}\par 


Existing works on network compression~\cite{lecun1990optimalbraindamamge, han2015nips, li2016l1filterprune, han2015deepcompress, Guo16nipsnetworksurgery} typically follow a common strategy to achieve compactness: pre-train a network and then iteratively prune and fine-tune the network to preserve benign accuracy. One of the most successful existing pruning strategies is weight-magnitude-based pruning~\cite{han2015nips, han2015deepcompress, li2016l1filterprune, frankleiclr19lotteryticket} where network parameters with the smallest weight magnitude are pruned. Multiple previous works~\cite{li2016l1filterprune, han2015nips, han2015deepcompress, Guo16nipsnetworksurgery} have demonstrated the success of this pruning strategy at preserving benign accuracy with very high compression ratios. However, except another concurrent research work~\cite{wijayanto2019towards}, none of these works studied the effect of such pruning strategies on the adversarial robustness of the pruned network. In this paper, we carefully study such interactions. \par 

Following pre-training, network pruning and fine-tuning are two key processes for creating compact networks, where former reduce network size while latter preserve performance after pruning. We first perform ablation studies where we investigate the impact of network pruning on robustness without any fine-tuning. Next, we investigate the integration of robust training mechanisms~\cite{madry2017towards, wang2018mixtrain} with both fine-tuning and pruning. However, as highlighted in multiple previous works~\cite{Zhoueccv16lessismoreeccv, leecilr19snip, frankleiclr19lotteryticket}, the use of the fine-tuning step incurs an additional computational overhead. 
The overhead is even higher with robust fine-tuning because it has significantly more complexity than natural training~\cite{madry2017towards}. Therefore, it is natural to ask whether fine-tuning with a less computationally intensive objective (such as fine-tuning with natural training) can successfully preserve robustness? In addition, the question of whether it is even feasible to design a compact and robust neural network by following a robust fine-tuning strategy is also an open one. 
\par 

Finally, even if the integration of robust training mechanisms with the full compression pipeline does result in robust networks, it will be further necessary to answer whether it has any advantage over robustly training a compact network from scratch? In this work, we answer these questions with the following contributions. \par

\noindent \textbf{Contributions.} (1) Our first contribution is to demonstrate that adversarial robustness of a network is almost as stable as benign accuracy, i.e., with increasing pruning ratios, the percentage decrease in both are similar in the absence of any fine-tuning step. 2) Following this, we show that due to the imbalance between training objective at pre-training and fine-tuning, when existing pruning and (benign) fine-tuning techniques are applied to a robustly trained network, the pruned network fails to provide robustness.  
Next, we show that when the fine-tuning objective is modified to be similar to the training step in adversarially robust or verifiably robust training, 
the network can maintain robustness 
up to compression ratios as large as 10$\times$. On average we are able to preserve up to 93\% benign accuracy, 92.5\% empirical robust accuracy, and 85.0\% verifiable robust accuracy for 10$\times$ compression ratio. (3) Finally, we demonstrate the advantage of the proposed compression pipeline over the baseline approach of training a compact network from scratch. While previous work has shown that pruning is advantageous in the context of benign network accuracy~\cite{li2016l1filterprune, han2015nips}, we show that in the context of empirical robust accuracy, the advantage over the baseline is even larger (up to 3.8\% and 10\% improvement in robust and benign accuracy, respectively).


\label{sec:background}

\section{Background and related work}
For image classification, the objective of deep neural networks is to learn an accurate classification of input images ($\mathcal{X}$) as true labels ($\mathcal{Y}$). This is often achieved by minimizing the empirical loss $\mathcal{L(\theta, X, Y)}$ over the training dataset, where $\theta$ refers to the network parameters.  \par

\noindent \textbf{Adversarial examples.}
Multiple previous works~\cite{goodfellow2014explaining, szegedy2013intriguing, carlini2017towards, athalye2018obfuscated} have demonstrated the success of adversarial examples, where an imperceptible perturbation ($\epsilon$) is added to the input at inference time. The goal of the adversary is to get this adversarial example misclassified. Multiple adversarial attacks have been proposed to generate a successful perturbation under different constraints~\cite{goodfellow2014explaining, carlini2017towards, madry2017towards, athalye2018robust}. We focus on attacks that utilize signed projected gradient descent~\cite{madry2017towards} to solve the underlying optimization problem for generating the perturbation.

\noindent \textbf{Adversarial robustness.} This line of research aims to 
search for network parameters that minimize $\mathcal{L(\theta, X, Y)}$ while maintaining robustness to 
adversarial attacks. This objective is often achieved by solving a robust optimization problem~\cite{madry2017towards, tramer2017ensembleadvtrain, zhang2019tradeoff, song2018robustdomainadapt, Shafahi2019freeadvtraining}, where the loss is minimized over both benign and adversarial inputs. It has been noted previously that training with stronger adversary often leads to higher robustness against adversarial attacks~\cite{tramer2017ensembleadvtrain, madry2017towards, zhang2019tradeoff}. However, since such robustness is only evaluated with gradient-based attacks, it is often referred as \emph{empirical robustness}.\par 

\noindent{\bf Verifiable robustness.} Another line of research called verifiably robust training~\cite{wang2018mixtrain,wong2018scaling,mirman2018differentiable,gowal2018effectiveness,raghunathan2018certified} aims to empower the neural networks to have \emph{verified robustness}. Verifiably robust training first uses sound verification techniques~\cite{reluval2018,kolter2017provable,shiqi2018efficient,raghunathan2018semidefinite,dvijotham2018dual,weng2018towards,zhang2018crown} to over-approximate the output of neural networks and then learns the parameters to minimize $\mathcal{L(\theta, X, Y)}$ with provable robustness properties. 
However, they struggle to scale up to large networks and datasets. In this paper, we use MixTrain, a recent approach by Wang et al.~\cite{wang2018mixtrain}, to obtain verifiably robust networks since it can scale up to moderately large networks. Note that our method is generic and thus can be used with any other verifiably robust training technique.


\noindent \textbf{Neural network Pruning.} The success of neural network pruning in achieving highly compact networks has been explored in multiple previous works~\cite{lecun1990optimalbraindamamge, han2015nips, han2015deepcompress, Zhoueccv16lessismoreeccv, Guo16nipsnetworksurgery, frankleiclr19lotteryticket, leecilr19snip, liu2018rethinking}. The key strategy in all these works is to first train a network followed by pruning and fine-tuning steps. The pruning process could be a single step or iterative pruning, followed by a fine-tuning process. \par 
Following the classification made by Liu et al.~\cite{liu2018rethinking}, pruning strategies can be divided into two categories: structured and unstructured pruning. In structured pruning, the objective is to prune multiple filters or neurons in convolutional and fully connected layer respectively. This makes the network compact by effectively reducing the size of each layer. In unstructured pruning, the objective is to only prune connection between neurons, which make the weight matrices in each layer sparse. 

\noindent \textbf{Robustness vs Compactness.} A closely related line of research direction focus on how network compression by itself impact the robustness of the network against adversarial attacks~\cite{wangSIP2018PruningDefense, guoNIPS18SparseDnn, DhillonICLR2018StochasticPruning, luo2019RandomMask, wangSIP2018PruningDefense, Zhao18shoudCompress}. In contrast, our works focus on achieving highly compressed networks while achieving robustness in parallel using state-of-the-art adversarial defenses. Another closely related work is Wijayanto et al.~\cite{wijayanto2019towards}, where the authors primarily focus on a specific compactness strategy called quantization. In contrast, our work focus on achieving compactness through network pruning. Additionally, our work considers much stronger attacks, both empirical and verified notions of robustness, and follows both structured and unstructured pruning strategies.


\section{Methodology} \label{sec:methodoloty}
\noindent \textbf{Formalizing network compression using pruning.} Designing compact neural networks using pruning is often a two-step process: network pre-training and fine-tuning. A typical optimization strategy for both steps is as follows:

\setlength{\abovedisplayskip}{3pt}
\setlength{\belowdisplayskip}{3pt}

\begin{minipage}{.5\textwidth}
\vspace{-10pt}
\noindent \textit{Step-1: Pre-training}
\begin{equation}\label{eq:prune}
    \hat{\theta}_p = argmin_{\theta} L_p(\theta, X, Y)
\end{equation}
\end{minipage}
\begin{minipage}{.5\textwidth}
\vspace{-5pt}
\centering
\noindent \textit{Step-2: Pruning with Fine-tuning}
\begin{align}
\hat{\theta}_f &= argmin_{\theta} L_f(\theta, X, Y) \label{eq:finetune_1} \\
s.t.~~~&1-\frac{||\theta||_0}{||\hat{\theta}_p||_0}\leq t
\label{eq:finetune_2}
\end{align}
\vspace{1pt}
\end{minipage}

Where $\theta, X, Y$ refers to network weight parameters, training images, and labels respectively. 
In the pre-training step, the network is trained following standard Empirical Risk Minimization (ERM)~\cite{vapnik1999overview,vapnik1992principles}. Specifically, the goal is to find the weights $\hat{\theta}_p$ that will minimize the pre-training loss objective $L_p$ over the training dataset.
In step-2, based on the pre-trained weights $\hat{\theta}_p$, the fine-tuning objective is to minimize another objective loss $L_f$ to maintain the network performance while also pruning network connections to achieve compactness. The objective for network pruning is modeled by adding a constraint on the sparsity of network parameters ($\theta$). In general the constraint is to prune a fraction ($t$) of the weight parameters obtained from pre-training ($\hat{\theta}_p$) i.e., $1-||\theta||_0/||\hat{\theta}_p||_0\leq t$.



To solve the optimization problem in step-2, the widely used approach is to solve the fine-tuning objective (Equation~\ref{eq:finetune_1}) and pruning constraint (Equation~\ref{eq:finetune_2}) alternatively. This requires multiple iterations, each pruning out a small fraction of network weights and minimizing the fine-tuning loss objective $L_f$ with pruned weights. For each iteration $j$, a pruning mask $M^j$ is first applied to weights $\theta^j$ achieving the newly pruned weights $M^j\odot\theta^j$, where $\odot$ refers to element-wise multiplication. Then $\theta^{j+1}$ is optimized as $argmin_{\theta} L_f(\theta, X, Y)$ with $\theta$ initialized with $M^j\odot\theta^j$. Such alternative iterations continue until the pruning ratio reaches the targeted threshold $t$.

Pruning mask $M$ has binary entries where a zero entry implies that the corresponding connection of the network is pruned. With structured pruning, instead of individual connections, individual filters and neurons from convolutional and fully connected layers are pruned respectively. Thus for structured pruning, the binary entries correspond to whether an individual filter/neuron is pruned. 



Next we discuss the methodology for our experiments in light of the aforementioned formalization.\par 

\noindent \textbf{Pruning without fine-tuning.} To preserve network performance, step-2 includes both pruning and fine-tuning processes. In the absence of network fine-tuning (i.e., we simply prune away a specific number of connections), one can characterize how robust current neural networks are to network pruning without any fine-tuning steps. With both structured and unstructured pruning, we will study the stability of network robustness with network pruning (without fine-tuning) and compare it with previously observed stability of benign accuracy~\cite{li2016l1filterprune, han2015nips}. \par 

\noindent \textbf{Balancing pre-training and fine-tuning objectives.} The generic formulation of network pruning allows minimizing a different loss objective in both pre-training and fine-tuning steps (Equation~\ref{eq:prune},~\ref{eq:finetune_1}). In previous work, where the challenge was to preserve benign accuracy, both $L_p, L_f$ are selected following natural training~\cite{han2015nips, han2015deepcompress, li2016l1filterprune, Guo16nipsnetworksurgery}. However, preserving robustness for compact networks requires integration with robust training. Our initial experiments investigated the effect of when fine-tuning loss objective ($L_f$) is selected without accounting for pre-training loss objective ($L_p$). We demonstrate the limitation of network pruning when there exist such imbalance between the two objectives. Following this, we evaluate the robustness of a compact network when both pre-training and fine-tuning loss objectives are similar to each other.  \par 
To satisfy constraint~\ref{eq:finetune_2}, we follow weight magnitude based pruning approach where the connections with smaller weight magnitude are pruned away. With structured pruning, this implies pruning the filter with the least $l_1$-norm. Previous works have highlighted the success of this heuristic in preserving network benign accuracy. We aim to demonstrate that such heuristic can maintain properties beyond benign accuracy, such as empirical and verifiable robustness. \par 

\noindent \textbf{Advantage over training compact networks from scratch.} It can be noted that the fine-tuning steps in the aforementioned formulation further adds additional computation cost after pre-training. When such cost is undesirable, it is natural to question whether network pruning has any advantage over training a compact network from scratch~\cite{madry2017towards}. We construct these baseline networks by removing an equal fraction of filter or neurons randomly from each layer and then robustly training from scratch. We compare this with compact networks achieved with structured pruning where the pruning algorithm decides which filter or neuron to remove. We present these results in Section~\ref{sec:why_pruning}.

\vspace{-10pt}
\label{sec:evaluation}
\vspace{-1pt}
\section{Experimental setup}
\vspace{-6pt}
In our experiments, we report results with both unstructured and structure pruning methods on the CIFAR-10 dataset. The key networks trained with these datasets are VGG16~\cite{simonyan2014VGG}, Wide-ResNet-28-5~\cite{zagoruyko2016WideResNet} CIFAR-large~\cite{kolter2017provable}, and All-convolutional network~\cite{springenberg2014AllCNN}. All networks are trained with SGD with starting learning rate of 0.1, with a step-wise decay by a factor of 10 after 50, 75, and 90\% of the training epochs. The momentum coefficient is 0.0001 along with a batch size of 128. We use the standard data augmentation procedures use in previous works for each dataset. The codebase is implemented in PyTorch~\cite{pytroch}. We will further make our code public for reproducibility of research. \par 
For adversarial attacks, we use signed projected gradient descent where maximum perturbations budget, per-step perturbation, and the number of iterations are 8, 2, and 10 respectively. This design choice is motivated by their success in previous works~\cite{madry2017towards, song2018robustdomainadapt}. 
We use similar attack parameters for adversarial training of the networks. Similar to previous works~\cite{madry2017towards, xie2018feature} we only include adversarial examples in training. For verifiably robust training, we use MixTrain with the best-reported hyperparameters in~\cite{wang2018mixtrain} i.e., $k$ and $alpha$ to be $10$ and $0.7$. Following previous works~\cite{shiqi2018efficient, wang2018mixtrain, kolter2017provable, wong2018scaling}, we choose the maximum perturbation budget of 2 for verifiably robust training. \par

For network pruning, we use the iterative pruning strategy. After pre-training, we use 40 pruning steps, each step followed by 5 fine-tuning epochs. The learning rate is selected as 0.001 and other hyper-parameters are kept the same as pre-training.  \par 

\noindent \textbf{Metrics:} Following the definitions in~\cite{wang2018mixtrain}, we use three different performance metrics: benign accuracy, empirical robust accuracy (\textit{era}), and verifiable robust accuracy (\textit{vra}). The benign accuracy refers to network accuracy with the non-modified inputs. \textit{Era} measures the percentage of test samples that are robust under gradient-based attacks (PGD attacks if not specified). \textit{vra} measures the fraction of test samples that are verified to be robust by network verification methods in~\cite{wang2018mixtrain,wong2018scaling}. 

\section{Experiments: understanding network pruning for robust networks}


\subsection{Impact of pruning without fine-tuning on robustly trained networks}
In this section, we present the impact of network pruning on performance when no fine-tuning process is used. It highlights how stable each performance metric is when a fraction of connections are pruned from the network. We report stability of each of these three metrics: Benign accuracy, \textit{era}, and \textit{vra}. \par
\begin{figure}[!tb]
    \centering
    \begin{subfigure}[t]{0.48\textwidth}
        \centering
        \includegraphics[width=\linewidth]{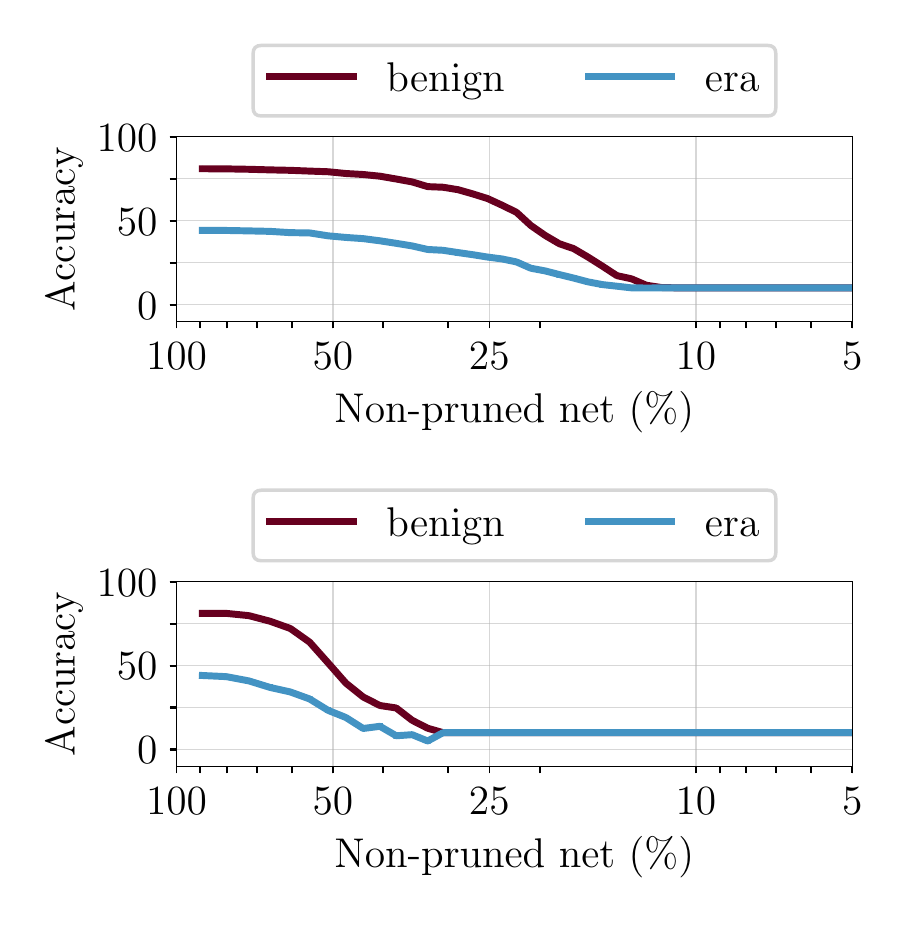}
        \caption{Adversarially trained \textit{VGG16} network.}
        \label{fig:fault_tolerance_era}
    \end{subfigure}
    \begin{subfigure}[t]{0.48\textwidth}
        \centering
        \includegraphics[width=\linewidth]{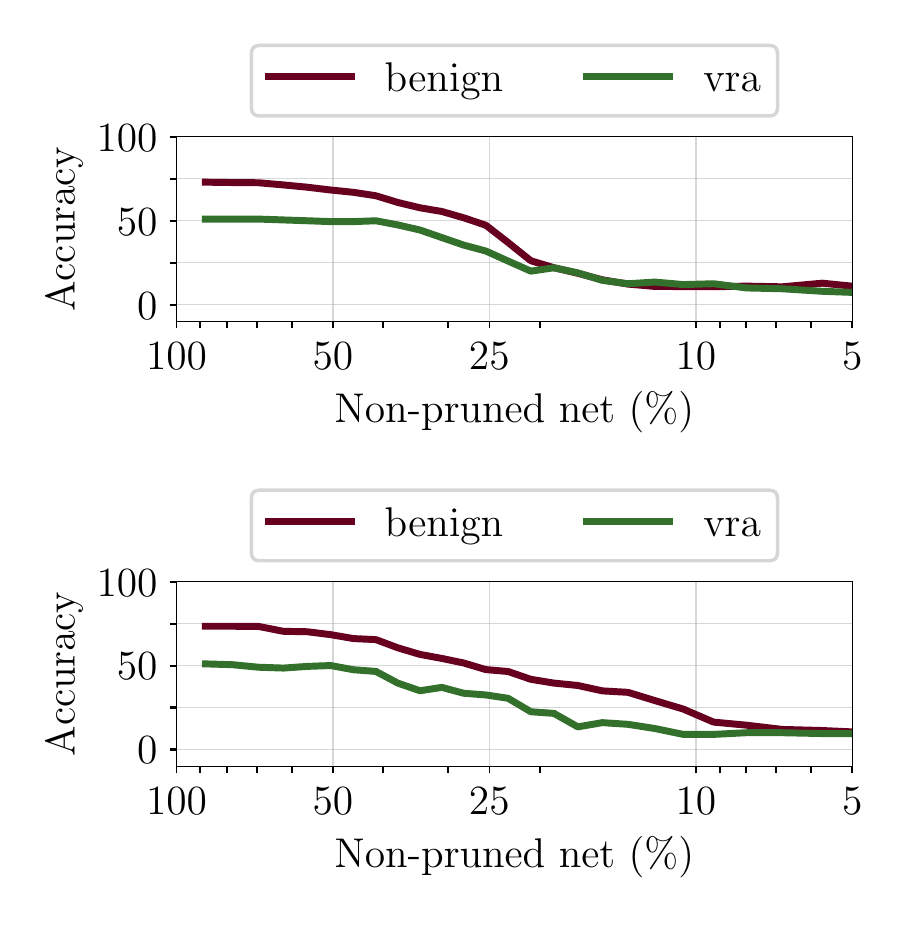}
        \caption{Verifiably trained \textit{CIFAR-large} network}
        \label{fig:fault_tolerance_vra}
    \end{subfigure}
    \caption{\small \it Degradation of key performance metrics with increasing pruning ratios for unstructured (top) and structured (bottom) pruning without any fine-tuning.}
    \label{fig:fault_tolerance}
    \vspace{-10pt}
\end{figure}

 
Fig.~\ref{fig:fault_tolerance} shows the stability of each of these metrics for both unstructured and structured pruning. \textit{We observe that the robustness of the network is almost as stable as its benign accuracy}. For example, for adversarially trained VGG16 network \textit{era} decreases by 5\% of its original value at 49\% and 28\% pruning ratios for unstructured and structured pruning respectively. In contrast, the benign accuracy decreases by 5\% of its original value at 59\% and 34\% pruning ratios respectively. Similarly, for CIFAR-large network, we observe that both benign accuracy and \textit{vra} decrease by 5\% at 49\% pruning ratio for both structured and unstructured pruning. 

\subsection{Imbalance between pruning and fine-tuning objectives}
\label{sec:imbalance}

In the previous section, we show that even though network robustness is stable at low pruning ratios, it sharply degrades at higher pruning ratios. To preserve robustness at higher pruning ratios, 
the pruned network should be fine-tuned. In this section, we demonstrate that an imbalance between network pre-training and fine-tuning loss objective will lead to poor robustness of the compact network. Following the formulation of designing a compact networks (Equation~\ref{eq:prune}, \ref{eq:finetune_1}), multiple choices for the loss objective $L_p, L_f$ exist. To achieve compact and robust networks, we select the network pre-training loss ($L_p$) from either of adversarial training or verifiably robust training. \par 

\begin{figure}[!htb]
    \vspace{-10pt}
    \begin{subfigure}[t]{0.33\textwidth}
        \centering \captionsetup{width=.9\linewidth}
        \includegraphics[width=\linewidth]{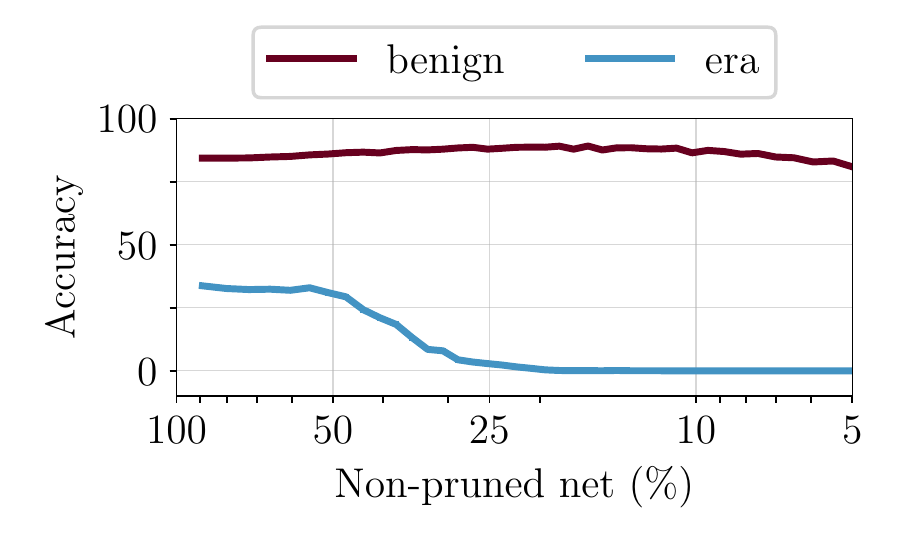}
        \caption{\small Fine-tuning with natural training an adversarially trained \textit{VGG16} network}
        \label{fig:imbalance_adv_trained}
    \end{subfigure}
    \begin{subfigure}[t]{0.32\textwidth}
        \centering \captionsetup{width=.9\linewidth}
        \includegraphics[width=\linewidth]{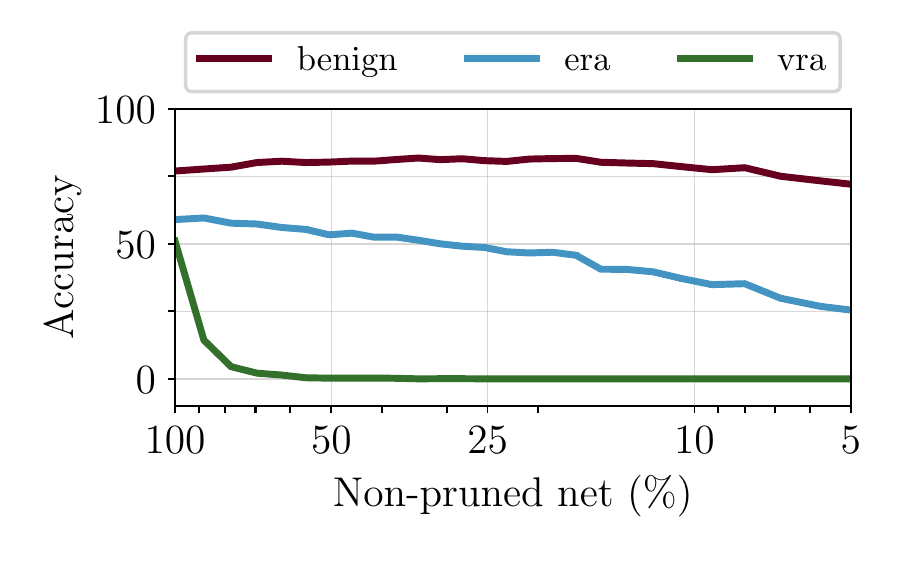}
        \caption{\small Fine-tuning with natural training on a verifiably trained \textit{CIFAR-large} network}
        \label{fig:imbalance_robust_trained1}
    \end{subfigure}
    \begin{subfigure}[t]{0.32\textwidth}
        \centering \captionsetup{width=.9\linewidth}
        \includegraphics[width=\linewidth]{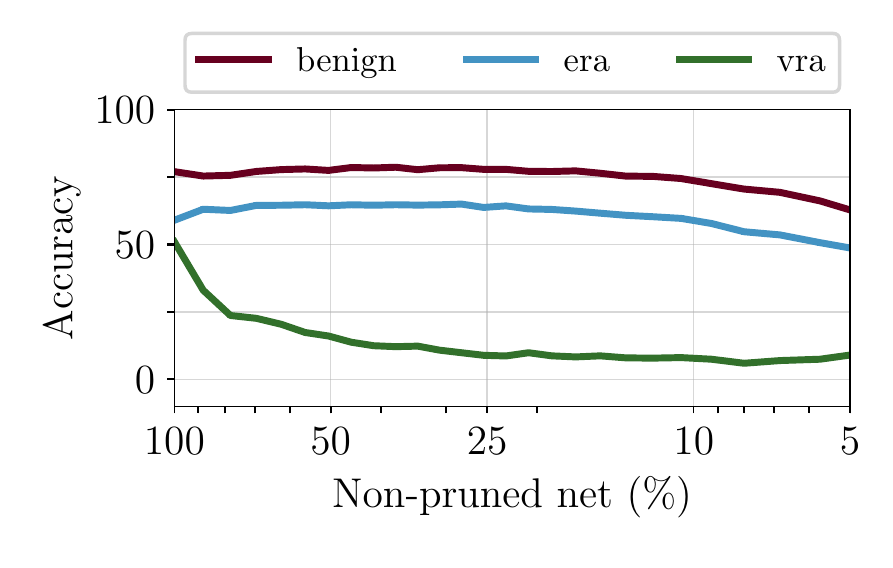}
        \caption{\small Fine-tuning with adversarial training on a verifiably trained \textit{CIFAR-large} network}
        \label{fig:imbalance_robust_trained2}
    \end{subfigure}
    \caption{\small \it Degradation of network robustness with increasing pruning ratios when fine-tuning objective is not balanced with training objective of the pre-trained network.}
    \label{fig:imbalanced_objectives}
\end{figure}

\begin{wraptable}{R}{0.6\linewidth}
\vspace{-10pt}
\caption{\small \it Fine-tuning with different $k$ in MixTrain for an network pre-trained with $k$=10. We report vra with different unstructured pruning ratios.}
\label{table:var_with_k}
\centering
\renewcommand{\arraystretch}{1.05}
\tabcolsep=0.11cm 
\resizebox{0.95\linewidth}{!}{
\begin{tabular}{llllllll}
\toprule
Pruning ratio (\%) & 50   & 72   & 84    & 90   & 92   & 95   & 98   \\ \midrule
MixTrain k=1        & 50.5 & 43.5 & 43.5  & 41.5 & 39.5 & 34.5 & 19.0   \\ 
MixTrain k=10       & 51.0 & 47.0 & 48.5  & 45.5 & 41.0 & 40.5 & 28.5 \\ \bottomrule 
\end{tabular}
}
\vspace{-10pt}
\end{wraptable}

Fig.~\ref{fig:imbalance_adv_trained} highlights the impact of imbalance between training objective for a network pre-trained using adversarial training~\cite{madry2017towards} and fine-tuned using natural training.
Though the benign accuracy of the network is preserved, \textit{era} decreases significantly with increasing pruning ratios. The effect could be attributed to the reason that natural training doesn't consider any adversary in its loss objective. Fine-tuning with it for a smaller number of epochs i.e., low pruning ratio, keeps the robustness intact. However, with the increasing number of epochs, the fine-tuning updates of network dominate as the network robustness quickly degrades to zero. We present the detailed results with different choices of $L_p$ and $L_f$ in Supplementary Section A, B. \par

Fig.~\ref{fig:imbalance_robust_trained1}, \ref{fig:imbalance_robust_trained2} report the results for a verifiably robust network fine-tuned with natural training and adversarial training respectively. In both experiments, \textit{vra} sharply decreases at very low pruning ratios. However, the properties aligned with fine-tuning objectives are preserved. For example, fine-tuning with natural and adversarial training are able to maintain the benign accuracy and \textit{era} respectively up to large pruning ratios but failed to preserve \textit{vra}. Even with fine-tuning using verifiably robust training, we show that difference in adversary's strength can lead to a large difference in robustness of the resulted compact network. For example, verifiably robust training with MixTrain~\cite{wang2018mixtrain} allows an increase in strength of adversary with parameter $k$ (the larger, the stronger robustness MixTrain can provide). Table~\ref{table:var_with_k} reports the results with $k= 1, 10$ where the network pre-training is performed with $k=10$. We can see that the difference in verifiable robustness for both adversaries gets more pronounced with an increase in pruning ratio. 
\par

\begin{wrapfigure}{R}{0.4\textwidth}
\vspace{-20pt}
    \centering
    \includegraphics[width=0.98\linewidth]{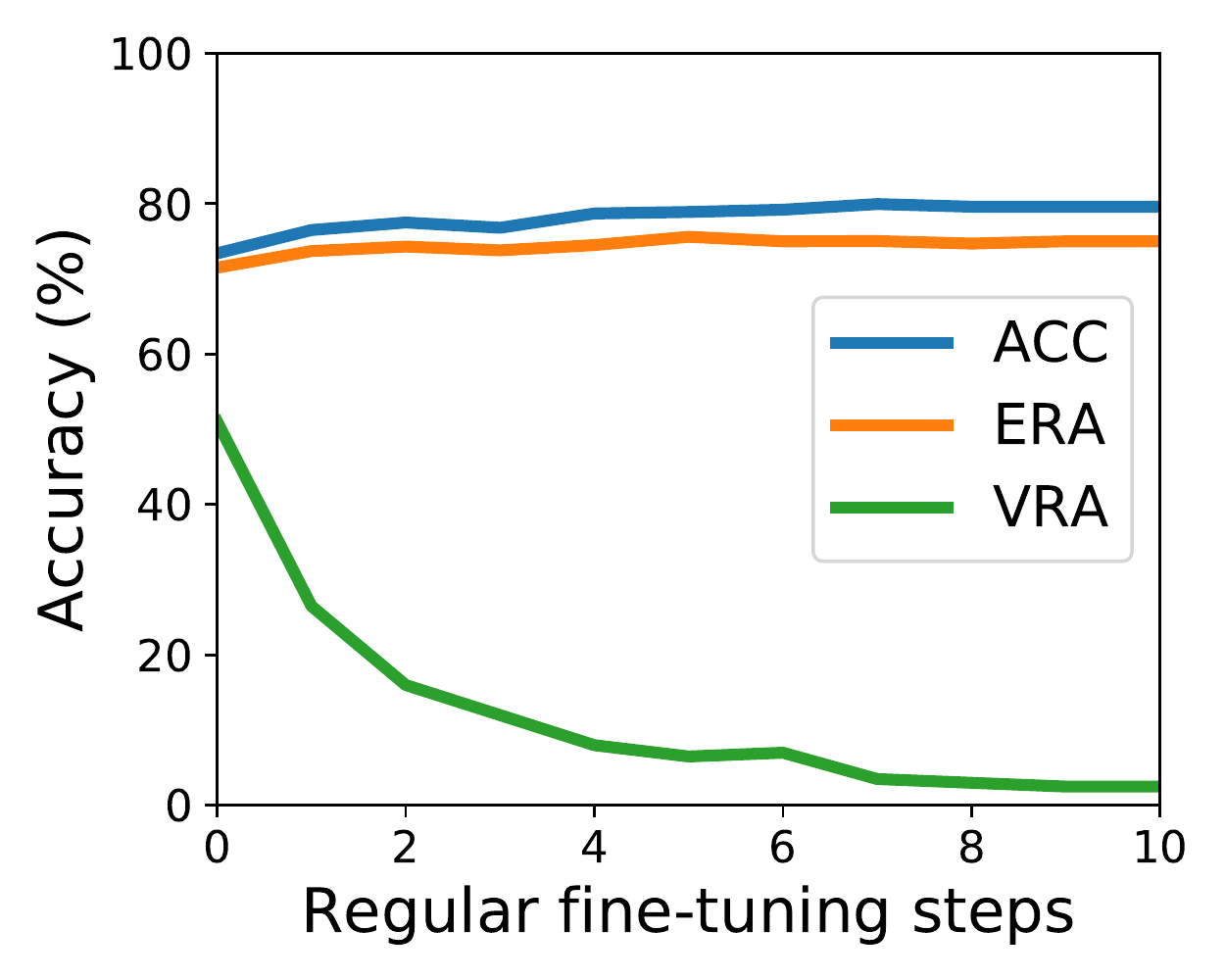}
    \caption{\small \it For verifiably trained CIFAR-large network, sharp decrease in \textit{vra} with a very small number of fine-tuning epoch using natural training without network pruning.}
    \label{fig:vra_one_epoch}
\vspace{-20pt}
\end{wrapfigure}

\noindent \textbf{Why \textit{vra} sharply degrades with natural fine-tuning?}
Since we follow an iterative pruning approach, Fig.~\ref{fig:imbalance_robust_trained1}, \ref{fig:imbalance_robust_trained2} show that fine-tuning the verifiable robust network for a few epochs with natural or adversarial training objective leads to significant loss of verifiable robustness. However, these results include both network pruning and fine-steps. In Figure~\ref{fig:vra_one_epoch}, we study how \textit{vra} changes with different epochs of fine-tuning with natural training in the absence of pruning. We can see that within only five epochs, \textit{vra} decrease from 51\% to 6.5\%. \par

\begin{wrapfigure}{R}{0.4\textwidth}
\centering
\vspace{-65pt}
\includegraphics[width=0.98\linewidth]{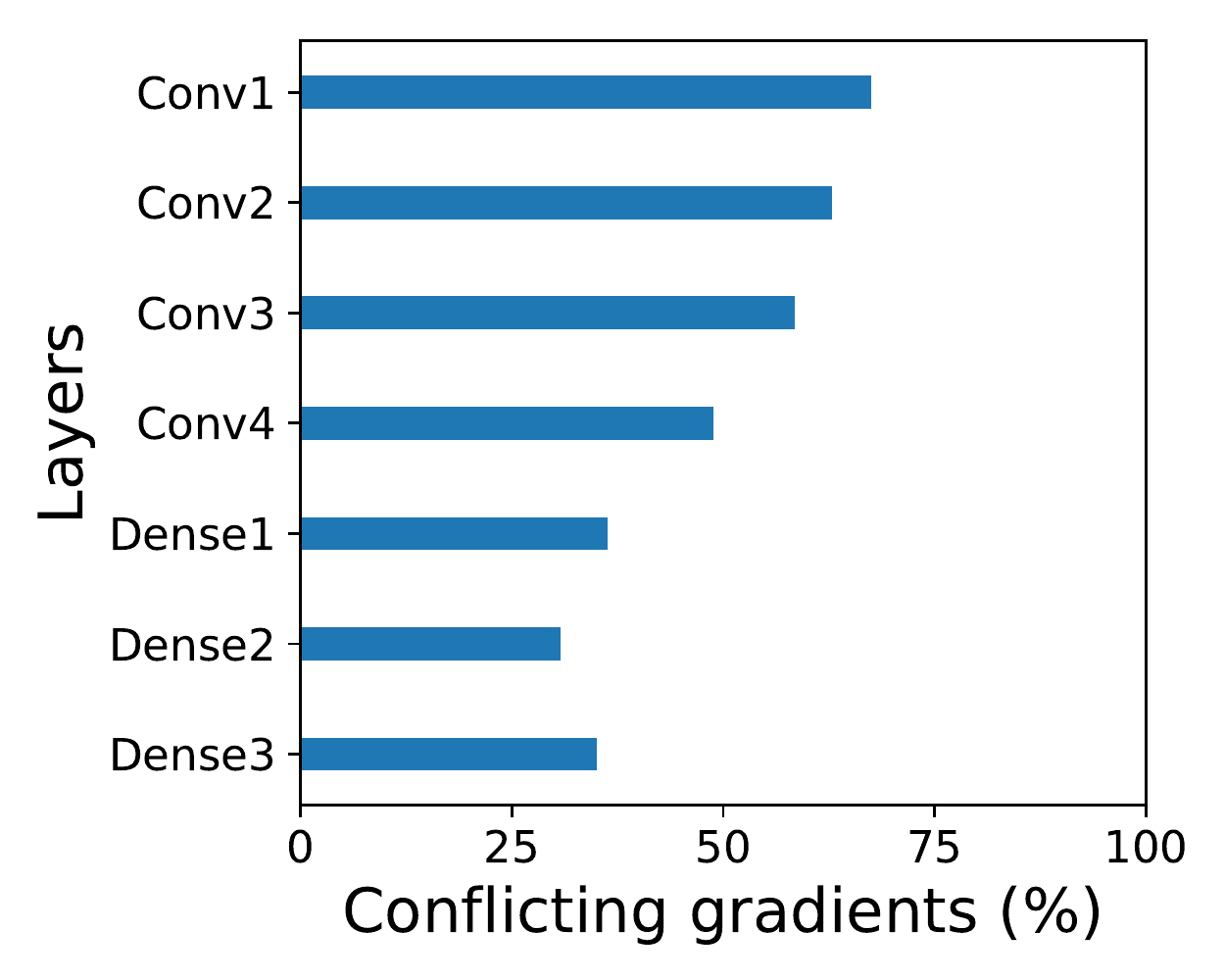}
\caption{\small \it Conflict in weight updates of natural and verifiably robust fine-tuning for an verifiably robust networks.}
\label{fig:weights_direction}
\vspace{-40pt}
\end{wrapfigure}

The natural question to ask now is whether the regular/benign loss objective truly conflicts with the verifiable robust loss objective? We take the same verifiably pre-trained robust network and measure the difference of gradients over the whole training set used to update its weights by verifiable robust fine-tuning and regular fine-tuning respectively. In Figure~\ref{fig:weights_direction}, we show on average 52\% of the weight gradients obtained by these two objectives are actually pointing to conflicting directions (e.g., verifiable robust loss points to positive direction while regular loss points to negative direction on updating one weight).

\section{Integrating compactness and robustness}
To preserve network robustness, we argue that it is critical to integrate the robust training objective in the network compression pipeline (Section~\ref{sec:methodoloty}). However, as highlighted in the previous section, using an imbalanced training objective in the network compression phase 
fails to preserve  robustness.
\textit{Now we demonstrate that selecting similar objectives in both pre-training and fine-tuning can successfully lead to robust and highly compact models}. 
 
\subsection{Preserving empirical robustness}
In this section, we demonstrate the success of fine-tuning based on adversarial training in preserving both benign accuracy and empirical robustness of the neural network with increasing pruning ratios. Each network in this experiment is pre-trained with adversarial training. The key results for unstructured and structured pruning are presented in Table~\ref{table:era_unstructured}, \ref{table:era_structured} respectively. We report results with three different fine-tuning strategies. First, we show the baseline results where no fine-tuning is performed (\textit{None}). Next, we capture the impact of fine-tuning based on natural training (\textit{Nat}). Finally, we demonstrate the success of fine-tuning with a similar objective as pre-training i.e., fine-tuning with adversarial training (\textit{Robust}).\par  

With both unstructured and structure pruning, we observe that \textit{Robust} fine-tuning leads to the best results for \textit{era}. For unstructured pruning, \textit{era} decreases only up to 3.5\% up for up to 10$\times$ compression ratio for VGG16 network. The improvement is even larger for unstructured pruning where the decrease in \textit{era} is only up to 1.7\% and 0.4\% for VGG16 and WRN-28-5 respectively. In addition to \textit{era}, the benign accuracy is also preserved to some extent with higher pruning ratios. For example, the decrease in benign accuracy is only 3.2\% for VGG16 at 10$\times$ compression ratios with unstructured pruning. Note that though \textit{Robust} pruning strategy preserves the network robustness, it also effectively suffers from the capacity vs. accuracy trade-off~\cite{madry2017towards} in adversarial training. This the reason why the network accuracy starts degrading to large extents when capacity is increased while robustness is preserved. We will discuss it in further detail in Section~\ref{sec:why_pruning}. \par 

We observe that the benign accuracy is comparatively difficult to preserve with structured pruning than unstructured pruning. With a compression ratio of 10$\times$, the benign accuracy is at least 9\% lower for structured pruning compared to unstructured pruning. However, the network \textit{era} is comparatively degraded less with structured pruning. We speculate that aggressive pruning ratios along with small decay in robustness could be the key factor in a large decrease in benign accuracy.
 
 \begin{table}[!htb]
 \vspace{-10pt}
\centering
\small
\renewcommand{\arraystretch}{1.05}
\caption{\small \it Summary of key results with \textbf{unstructured} pruning approach on adversarially trained networks.}
\label{table:era_unstructured}
\resizebox{0.9\linewidth}{!}{
\begin{tabular}{cccccccccc}

\toprule
\multirow{2}{*}{Network} & \multirow{2}{*}{\begin{tabular}[c]{@{}c@{}}Pruning \\
ratio (\%)\end{tabular}} & \multirow{2}{*}{\begin{tabular}[c]{@{}c@{}}ERA* (\%)\\ \end{tabular}} & \multicolumn{3}{c}{ERA (\%)} & \multirow{2}{*}{\begin{tabular}[c]{@{}c@{}}ACC* (\%)\\ \end{tabular}} & \multicolumn{3}{c}{ACC (\%)} \\ \cline{4-6} \cline{8-10}
& & & \multicolumn{1}{c}{None} & \multicolumn{1}{c}{Nat} & \multicolumn{1}{c}{Robust} & & \multicolumn{1}{c}{None} & \multicolumn{1}{c}{Nat} & \multicolumn{1}{c}{Robust} \\ \midrule
\multirow{3}{*}{VGG16} & 50 & \multirow{3}{*}{43.3}  & 40.0 & 31.8 & \textbf{41.4}  & \multirow{3}{*}{81.6} & 78.1 & 84.9 & 80.9 \\
 & 75 &  & 28.4 & 32.9 & \textbf{40.8}  &  & 63.2 & 85.9 & 80.2 \\
  & 90 &  & 10.0 & 22.9 & \textbf{39.8}  &  & 10.0 & 86.1 & 78.4 \\ \midrule
\multirow{3}{*}{\begin{tabular}[c]{@{}c@{}}Wide-Resnet\\ 28-5\end{tabular}} & 50 & \multirow{3}{*}{42.6} & 33.4 & 36.4 & \textbf{41.1} & \multirow{3}{*}{85.1} & 80.9 & 86.6 & 85.0 \\
& 75 &  & 16.6 & 28.6 & \textbf{39.8} &   & 56.4 & 87.0 & 84.7 \\
& 90 &  &  7.8 & 12.1 & \textbf{39.7} &   & 10.2 & 88.1 & 84.5 \\ \bottomrule

\multicolumn{10}{l}{$*$ denotes the \textit{era} and benign accuracy of pre-trained robust networks.} \\ 

\end{tabular}
}
\vspace{-10pt}
\end{table}

\begin{table}[!htb]
\centering
\small
\caption{\small \it Summary of key results with \textbf{structured} pruning approach on adversarially trained networks.}
\label{table:era_structured}
\renewcommand{\arraystretch}{1.05}
\resizebox{0.9\linewidth}{!}{
\begin{tabular}{cccccccccc}

\toprule
\multirow{2}{*}{Network} & \multirow{2}{*}{\begin{tabular}[c]{@{}c@{}}Pruning \\
ratio (\%)\end{tabular}} & \multirow{2}{*}{\begin{tabular}[c]{@{}c@{}}ERA* (\%)\\ \end{tabular}} & \multicolumn{3}{c}{ERA (\%)} & \multirow{2}{*}{\begin{tabular}[c]{@{}c@{}}ACC* (\%)\\ \end{tabular}} & \multicolumn{3}{c}{ACC (\%)} \\ \cline{4-6} \cline{8-10}
& & & \multicolumn{1}{c}{None} & \multicolumn{1}{c}{Nat} & \multicolumn{1}{c}{Robust} & & \multicolumn{1}{c}{None} & \multicolumn{1}{c}{Nat} & \multicolumn{1}{c}{Robust} \\ \midrule
\multirow{3}{*}{VGG16} & 50 & \multirow{3}{*}{43.3}  & 19.0 & 29.3 & \textbf{45.3} & \multirow{3}{*}{81.6} & 39.5 & 86.5 & 80.5 \\
& 75 &  & 10.0 & 2.8 & \textbf{47.6} &  & 10.0 & 87.9 & 78.5\\
& 90 &  & 10.0 & 0.0 & \textbf{41.6 } &  & 10.0 & 86.4 & 69.0\\ \midrule

\multirow{3}{*}{\begin{tabular}[c]{@{}c@{}}Wide-Resnet\\ 28-5\end{tabular}} & 50 & \multirow{3}{*}{42.6} & 10.0 & 4.2 & \textbf{44.5} & \multirow{3}{*}{85.1} & 14.7 & 89.8 & 82.3 \\
& 75 &  & 10.0 & 0.0 & \textbf{43.6} &  & 10.0 & 89.7 & 76.1 \\ 
& 90 &  & 10.0 & 0.0 & \textbf{34.3} &  & 10.0 & 87.6 & 64.4 \\ \bottomrule
\multicolumn{10}{l}{$*$ denotes the \textit{era} and benign accuracy of pre-trained robust networks.} \\ 

\end{tabular}
}
\vspace{-10pt}
\end{table}

\subsection{Preserving verifiable robustness}
In this section, we demonstrate the success of fine-tuning based on verifiably robust training with \textit{MixTrain} in preserving the benign accuracy and \textit{vra} with increasing pruning ratios. The networks evaluated in this section are pre-trained using \textit{MixTrain}. Similar to previous section we report the two baselines metric, \textit{None} and \textit{Nat}. Fine-tuning based on verifiably robust training is referred as \textit{Robust}. \par 

The key results for unstructured and structured pruning are presented in Table~\ref{table:vra_unstructured}, \ref{table:vra_structured} respectively.
These results highlight that fine-tuning with \textit{MixTrain} is highly successful in preserving \textit{vra}. As shown in Table~\ref{table:vra_unstructured} for unstructured pruning, robust fine-tuning only decreases by 4.5\% from the original \textit{vra} with compression ratio of 10$\times$. The decrease in benign accuracy is also comparatively small for unstructured pruning. For structured pruning, \textit{vra} and benign accuracy drop to $35.5\%$ and $59.5\%$ from $51\%$ and $73.4\%$, but it is still a relatively small decrease, compared to other fine-tuning strategies. \par

We experiment with different network architectures and observe a similar trend. With \textit{all-convolutional (ALL-CNN)} network, though the performance of the pre-trained network is relatively poor compared to \textit{CIFAR-large} network, the benign accuracy and \textit{vra} don't degrade to a large extent for very high pruning ratios. We omit results for 90\% pruning ratio as all filters get pruned out in some layers at this pruning ratio.

\begin{table}[!ht]
\vspace{-10pt}
\centering
\small
\caption{\small \it Key results for \textbf{unstructured} pruning approach on verifiably robust networks trained with MixTrain..}
\label{table:vra_unstructured}
\renewcommand{\arraystretch}{1.05}
\resizebox{0.9\linewidth}{!}{
\begin{tabular}{cccccccccc}

\toprule
\multirow{2}{*}{Network} & \multirow{2}{*}{\begin{tabular}[c]{@{}c@{}}Pruning \\
ratio (\%)\end{tabular}} & \multirow{2}{*}{\begin{tabular}[c]{@{}c@{}}VRA*(\%)\\ \end{tabular}} & \multicolumn{3}{c}{VRA  (\%)} & \multirow{2}{*}{\begin{tabular}[c]{@{}c@{}}ACC*(\%)\\\end{tabular}} & \multicolumn{3}{c}{ACC (\%)} \\ \cline{4-6} \cline{8-10}
& & & \multicolumn{1}{c}{None} & \multicolumn{1}{c}{Nat} & \multicolumn{1}{c}{Robust} & & \multicolumn{1}{c}{None} & \multicolumn{1}{c}{Nat} & \multicolumn{1}{c}{Robust} \\ \midrule
\multirow{3}{*}{\textit{CIFAR-large}} & 50 & \multirow{3}{*}{51.0} & 49.5 & 6.5 & \textbf{51.0} & \multirow{3}{*}{73.4} & 68.3 & 80.3 & 73.3 \\ 
& 75 & & 34.2 & 2.1 & \textbf{48.6} & & 47.4 & 82.0 & 73.2 \\  
& 90 & & 13.5 & 2.0 & \textbf{45.5} & & 10.8 & 80.9 & 66.9 \\   \midrule

\multirow{3}{*}{\textit{CIFAR-ALL-CNN}} & 50 & \multirow{3}{*}{42.3} & 42.5 & 1.4 & \textbf{34.1} & \multirow{3}{*}{60.0} & 56.5 & 66.0 & 55.6 \\ 
& 75 &  & 32.7 & 0.0 & \textbf{35.3} &  & 36.1 & 70.7 & 53.7 \\
& 90 &  & 17.0 & 0.0 & \textbf{34.2} &  & 16.8 & 68.9 & 51.2 \\ \bottomrule
\multicolumn{10}{l}{$*$ denotes the \textit{vra} and benign accuracy of pre-trained robust networks.}
\end{tabular}
}
\vspace{-10pt}
\end{table}




\begin{table}[!ht]
\centering
\small
\caption{\small \it Key results for \textbf{structured} pruning approach on verifiably robust networks trained with MixTrain.}
\label{table:vra_structured}
\renewcommand{\arraystretch}{1.05}
\resizebox{0.9\linewidth}{!}{
\begin{tabular}{cccccccccc}
\toprule
\multirow{2}{*}{Network} & \multirow{2}{*}{\begin{tabular}[c]{@{}c@{}}Pruning \\
ratio (\%)\end{tabular}} & \multirow{2}{*}{\begin{tabular}[c]{@{}c@{}}VRA* (\%)\\ \end{tabular}} & \multicolumn{3}{c}{VRA (\%)} & \multirow{2}{*}{\begin{tabular}[c]{@{}c@{}}ACC* (\%)\\ \end{tabular}} & \multicolumn{3}{c}{ACC (\%)} \\ \cline{4-6} \cline{8-10}
& & & \multicolumn{1}{c}{None} & \multicolumn{1}{c}{Nat} & \multicolumn{1}{c}{Robust} & & \multicolumn{1}{c}{None} & \multicolumn{1}{c}{Nat} & \multicolumn{1}{c}{Robust} \\ \midrule
\multirow{3}{*}{\textit{CIFAR-large}} & 50 & \multirow{3}{*}{51.0} & 46.5 & 3.5 & \textbf{47.5} & \multirow{3}{*}{73.4} & 68.4 & 80.2 & 72.7 \\ 
& 75 & & 32.5 & 0.0 & \textbf{44.5} & & 47.6 & 80.2 & 70.5 \\  
& 90 & & 9.0 & 0.0 & \textbf{35.5} & & 24.1 & 75.5 & 59.5 \\   \midrule

\multirow{2}{*}{\textit{CIFAR-ALL-CNN}} & 50 & \multirow{2}{*}{42.3} & 11.9 & 0.6 & \textbf{35.6} & \multirow{2}{*}{60.0} & 17.7 & 67.5 & 54.2 \\ 
& 75 &  & 9.9 & 0.0 & \textbf{23.5} &  & 10.0 & 50.7 & 36.8 \\ \bottomrule
\multicolumn{10}{l}{$*$ denotes the \textit{vra} and benign accuracy of pre-trained robust networks.} \\ 

\end{tabular}
}
\vspace{-10pt}
\end{table}

\section{Advantage of pruning to achieve compact networks}\label{sec:why_pruning}

\begin{wraptable}{R}{0.6\linewidth}
\vspace{-13pt}
\caption{\small \it Comparison of two distinct approaches to obtain compact models. First one is the methodology proposed in this paper (\textit{fine-tuned}) while in the second method a compact model is trained from scratch (\textit{scratch}). The current results are for \textit{structured} pruning approach for VGG16 network.}
\label{table:scratch_adv}
\centering
\small
\resizebox{0.95\linewidth}{!}{
\begin{tabular}{ccccccc}
\toprule
\multirow{2}{*}{\begin{tabular}[c]{@{}c@{}}Pruning \\ ratio (\%)\end{tabular}} &  & \multicolumn{2}{c}{Benign accuracy (\%)} &  & \multicolumn{2}{c}{\begin{tabular}[c]{@{}c@{}}\textit{ERA (\%)}\\ \end{tabular}} \\ \cline{3-4} \cline{6-7}
 &  & scratch & fine-tuned &  & scratch & fine-tuned \\ \midrule
 50 &  & 78.3 & \textbf{80.5} &  & 44.9 & \textbf{45.3 }\\
 75 &  & 72.1 & \textbf{75.5} &  & 42.7 & \textbf{47.6} \\
 90 &  & 59.2 & \textbf{69.0} &  & 37.8 & \textbf{41.6} \\ \bottomrule
\end{tabular}
}
\vspace{-10pt}
\end{wraptable}
In this section we quantify the benefits of network compression pipeline to designing compact networks by comparing it with compact models which are trained from scratch i.e., no pre-training and iterative fine-tuning strategy are used. With the baseline approach i.e., compact network trained from scratch, we observe that with large pruning ratios, the benign accuracy and robustness are significantly lower than the original network (Table~\ref{table:scratch_adv}). This highlights the trade-off curve which is observed in previous works~\cite{madry2017towards}. However, we show that following the approach of network pruning and fine-tuning the impact of the trade-off can be significantly reduced. For example, the proposed approach is able to improve benign accuracy by approximately 10\% when the network capacity is reduced by a factor of 10$\times$. Similarly, \textit{era} of the network is improved by 3.8\% over the baseline. \par 

\textit{We also observe that pruning is advantageous over training from scratch, but the gains are much higher if robustness also need to be preserved along with benign accuracy.} For example, if the objective is to only maintain benign accuracy, the relative benefits in accuracy over the baseline approach is only 4.2\% for 10$\times$ decrease in capacity (detailed results in Table 7 in supplementary material). However, with a strictly harder objective i.e., when both benign accuracy and robustness needs to be preserved, the proposed methods bring as much as 10\% improvement in benign accuracy over the baselines.  


\label{sec:discussion}


\label{sec:conclusion}

\section{Conclusion}
In this paper, we have extensively studied the relationship between neural network compactness and robustness. First, we observe that existing network pruning methods fail to achieve both compact and robust networks. We provide extensive insights into the limitations of previous works. To conquer these limitations, we propose improved network pruning methods which can successfully preserve the robustness of the networks to large extents for compression ratios as high as 10$\times$. In particular, our approach can preserve 93\% of benign accuracy, 92.5\% of empirical robust accuracy, and 85\% of verifiable robust accuracy for 10$\%$ compression ratio with structured pruning. The gains with structured pruning are 84.3\%, 90.4\%, and 73.8\% respectively. Finally, we demonstrate the advantage of proposed techniques over robust training of compact networks from scratch.  

\section{Acknowledgement}
This work was supported in part by the National Science Foundation under grants CNS-1553437, CNS-1704105, CIF-1617286 and EARS1642962, by the Office of Naval Research Young Investigator Award, by the Army Research Office Young Investigator Prize, by Faculty research awards from Intel and IBM.

\small

\bibliographystyle{abbrv}
\bibliography{ref}

\appendix

\section{Detailed results for empirical robustness (\textit{era}) relationship with network pruning} \label{appsec:imbalance_detailed_results}
\begin{figure*}[!htb]
    \centering
    \begin{subfigure}[t]{0.32\textwidth}
        \centering
        \includegraphics[width=\linewidth]{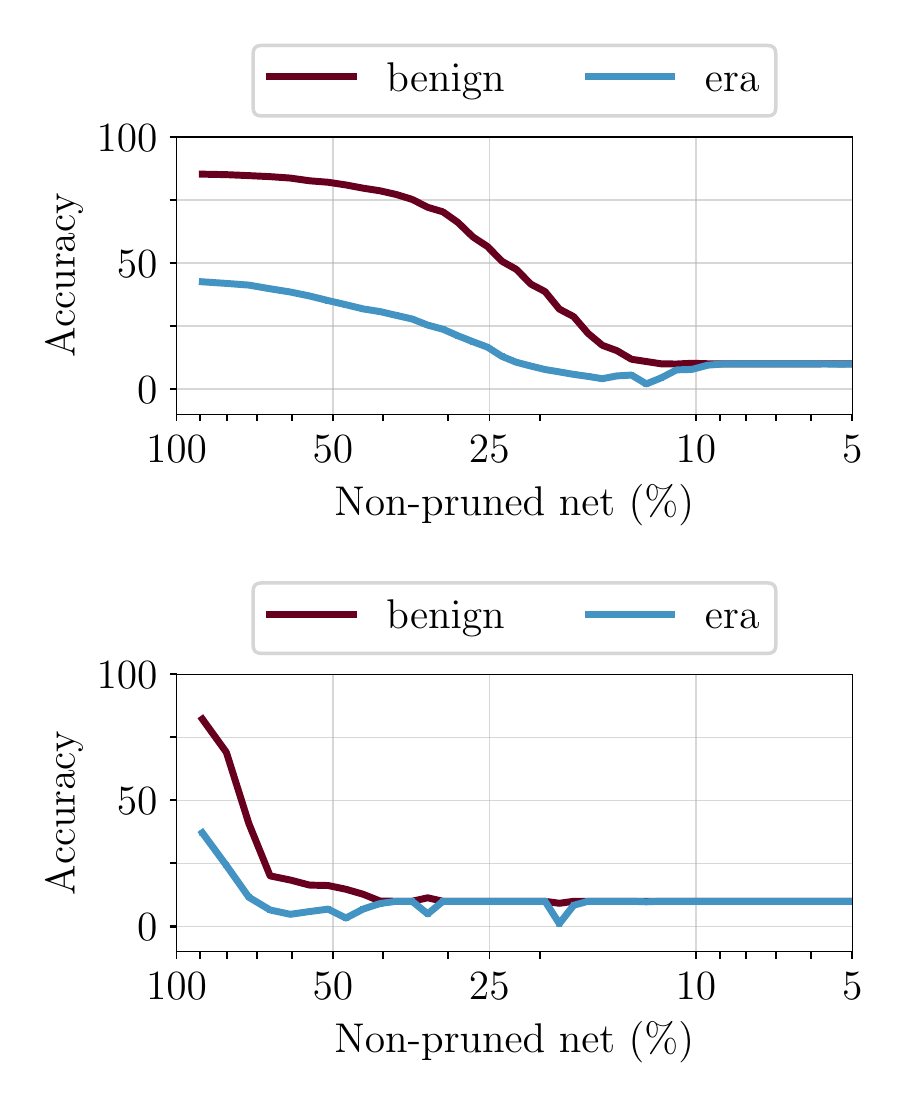}
        \caption{No fine-tuning.}
    \end{subfigure}
    \begin{subfigure}[t]{0.32\textwidth}
        \centering
        \includegraphics[width=\linewidth]{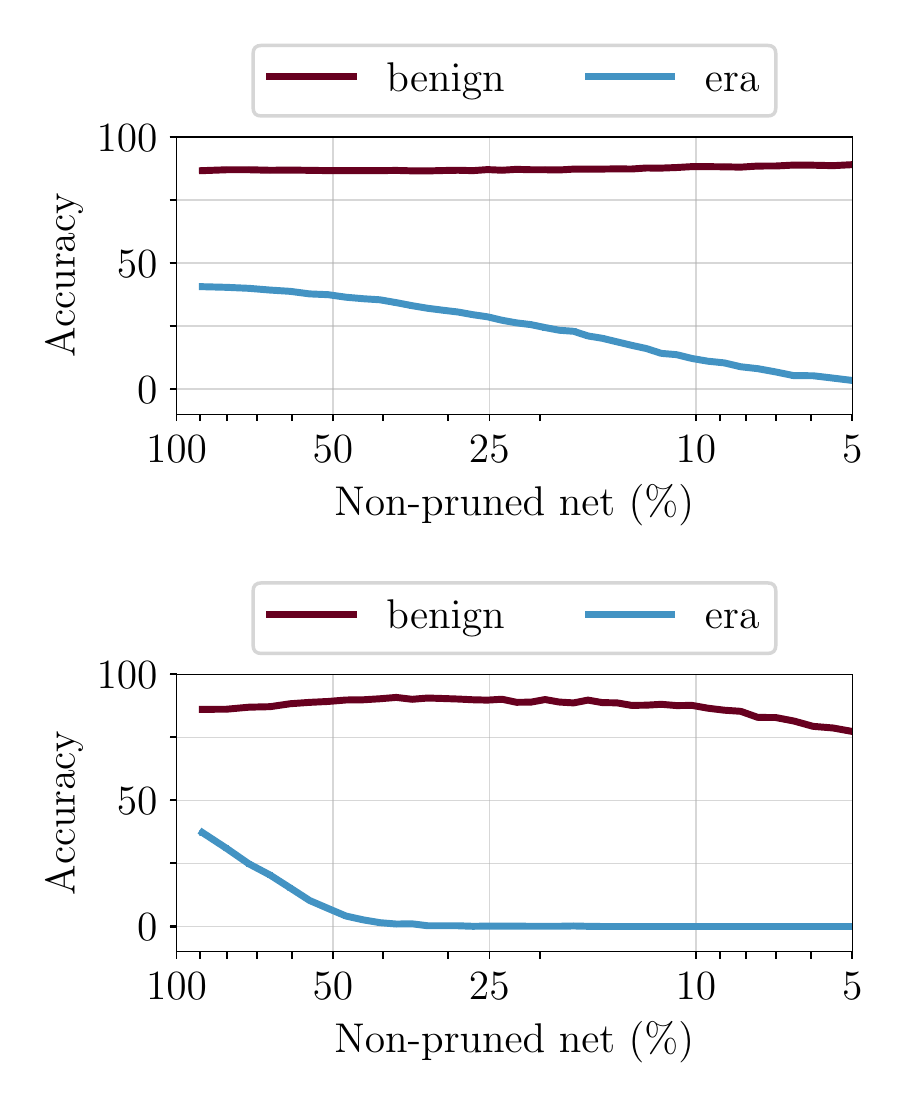}
        \caption{Fine-tuning with natural training}
    \end{subfigure} 
    \begin{subfigure}[t]{0.32\textwidth}
        \centering
        \includegraphics[width=\linewidth]{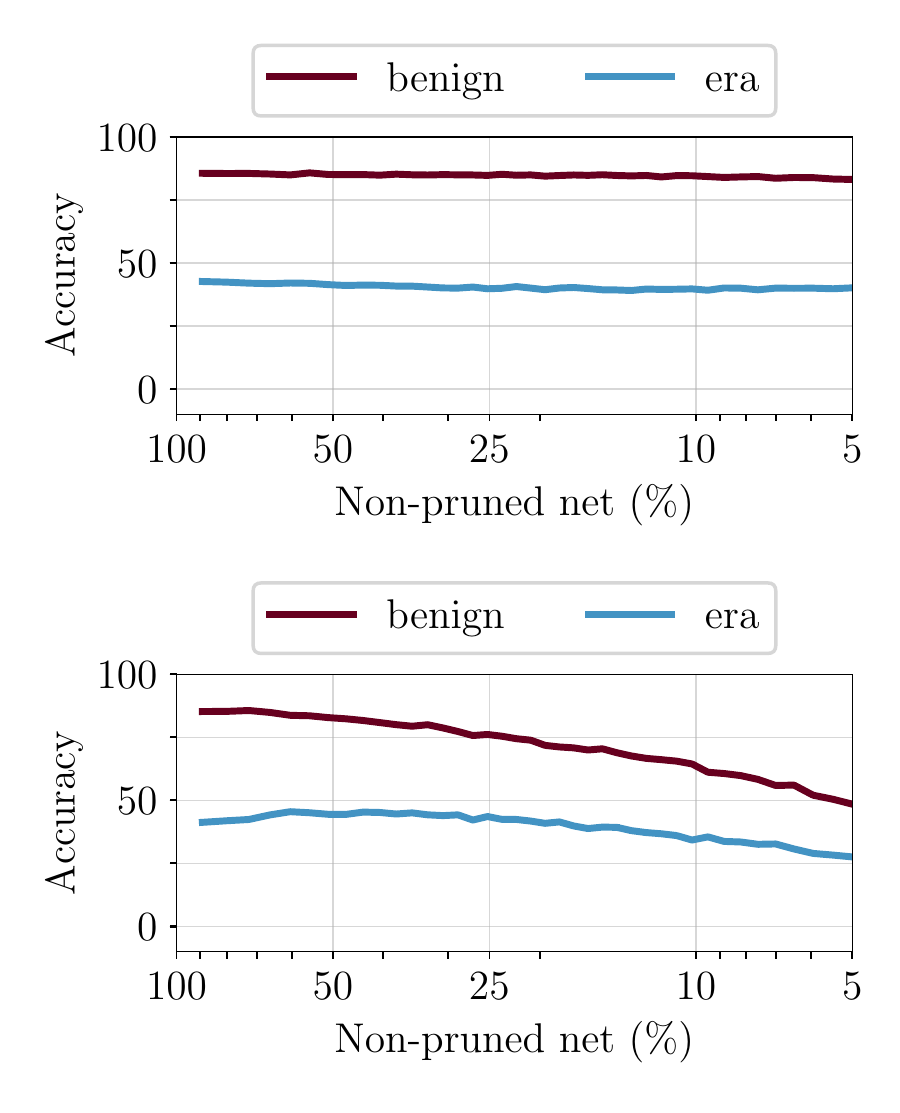}
        \caption{Fine-tuning with adversarial training}
    \end{subfigure}
    \caption{Summary of key results for WRN-28-5. The model is pre-trained with adversarial training.}
    \label{fig:adv_finetune_wrn_full}
\end{figure*}

In this section, we present detailed experimental study for different choices of pruning and fine-tuning loss objectives in the network compression pipeline. We first experiment with no fine-tuning strategy in section~\ref{appsubsec:no_finetune} and later present results for fine-tuning with natural and adversarial training in section~\ref{appsubsec:nat_finetune} and \ref{appsubsec:adv_finetune} respectively. Within each section, we experiment with VGG-16 network trained with both natural and adversarial training. We present results for WRN-28-10 in Figure~\ref{fig:adv_finetune_wrn_full}.
\subsection{No Fine-tuning} \label{appsubsec:no_finetune}
\begin{figure*}[!htb]
    \centering
    \begin{subfigure}[t]{0.48\textwidth}
        \centering
        \includegraphics[width=\linewidth]{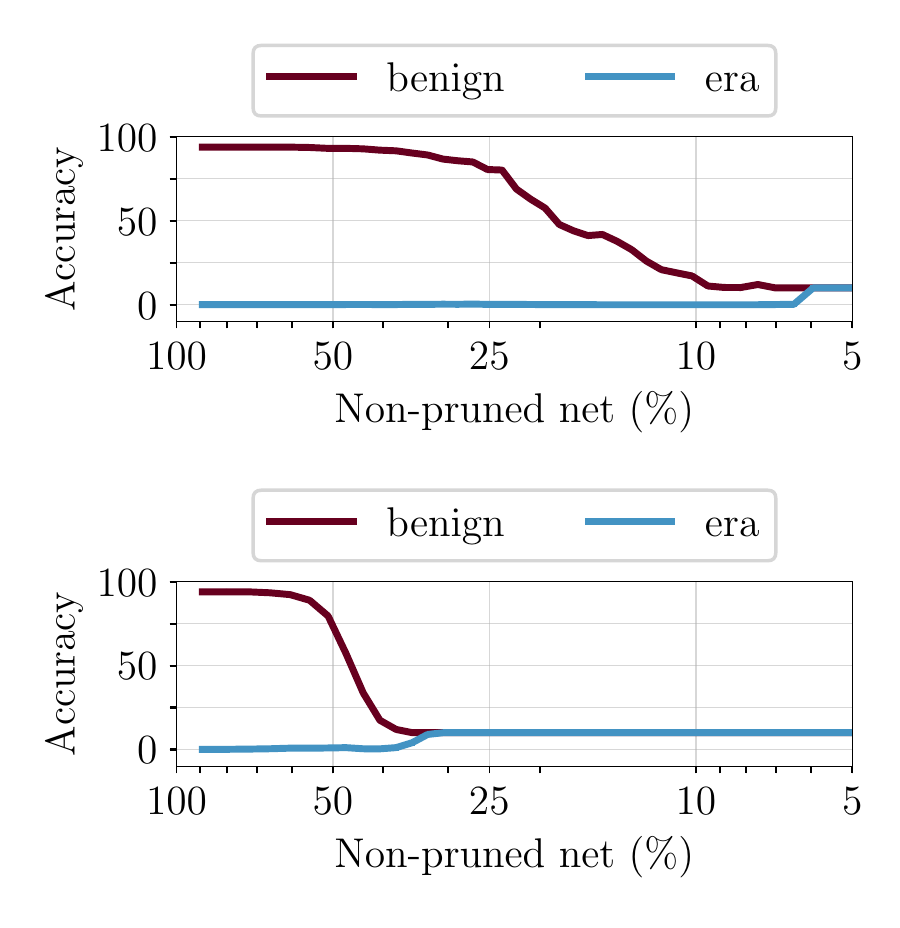}
        \caption{Naturally trained model}
    \end{subfigure}
    \begin{subfigure}[t]{0.48\textwidth}
        \centering
        \includegraphics[width=\linewidth]{images/no_finetune_vgg_b.pdf}
        \caption{Adversarially trained model}
    \end{subfigure} 
    \caption{No fine-tuning for VGG16 network trained on CIFAR-10 dataset. The top plot shows results for unstructured pruning while the bottom one captures results for structured pruning based approach.}
    \label{fig:no_finetune_vgg_full}
\end{figure*}

\subsection{Fine-tuning based on natural  training.}\label{appsubsec:nat_finetune}
\begin{figure*}[!htb]
    \centering
    \begin{subfigure}[t]{0.48\textwidth}
        \centering
        \includegraphics[width=\linewidth]{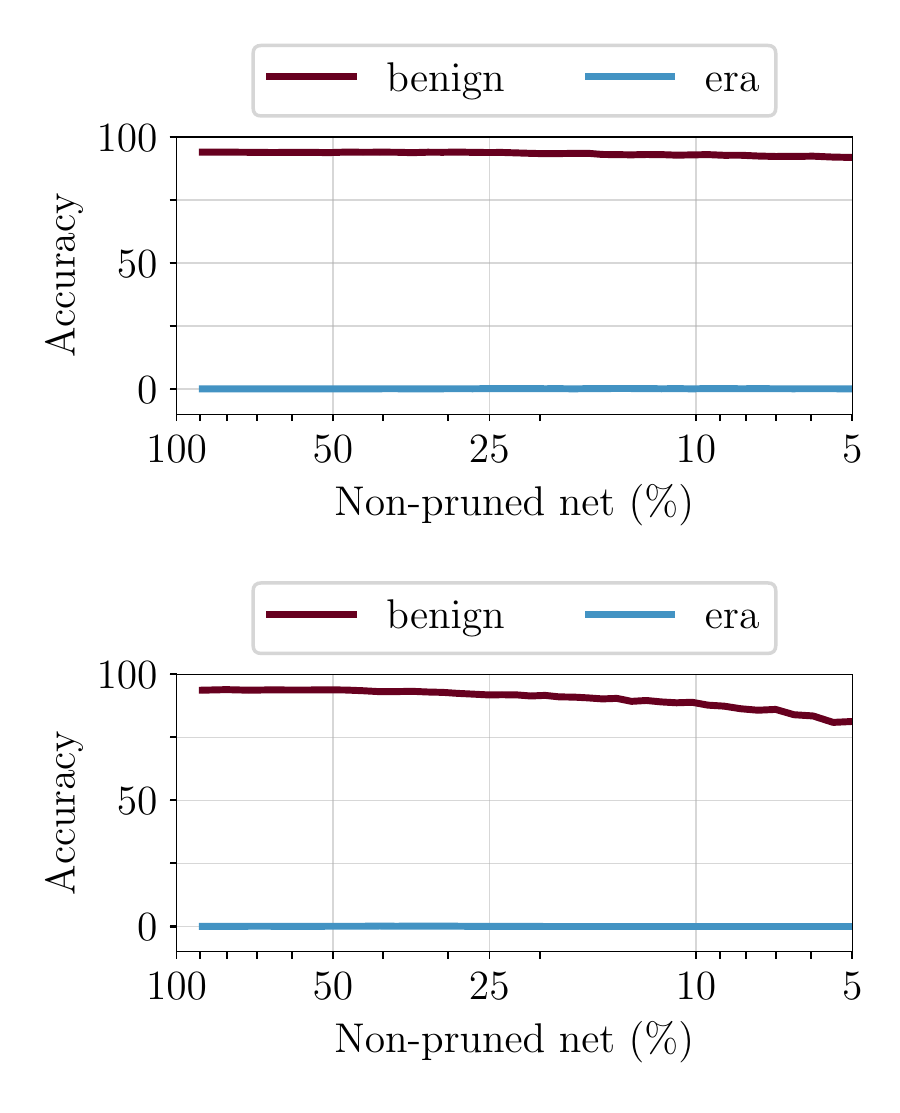}
        \caption{Naturally trained model}
    \end{subfigure}
    \begin{subfigure}[t]{0.48\textwidth}
        \centering
        \includegraphics[width=\linewidth]{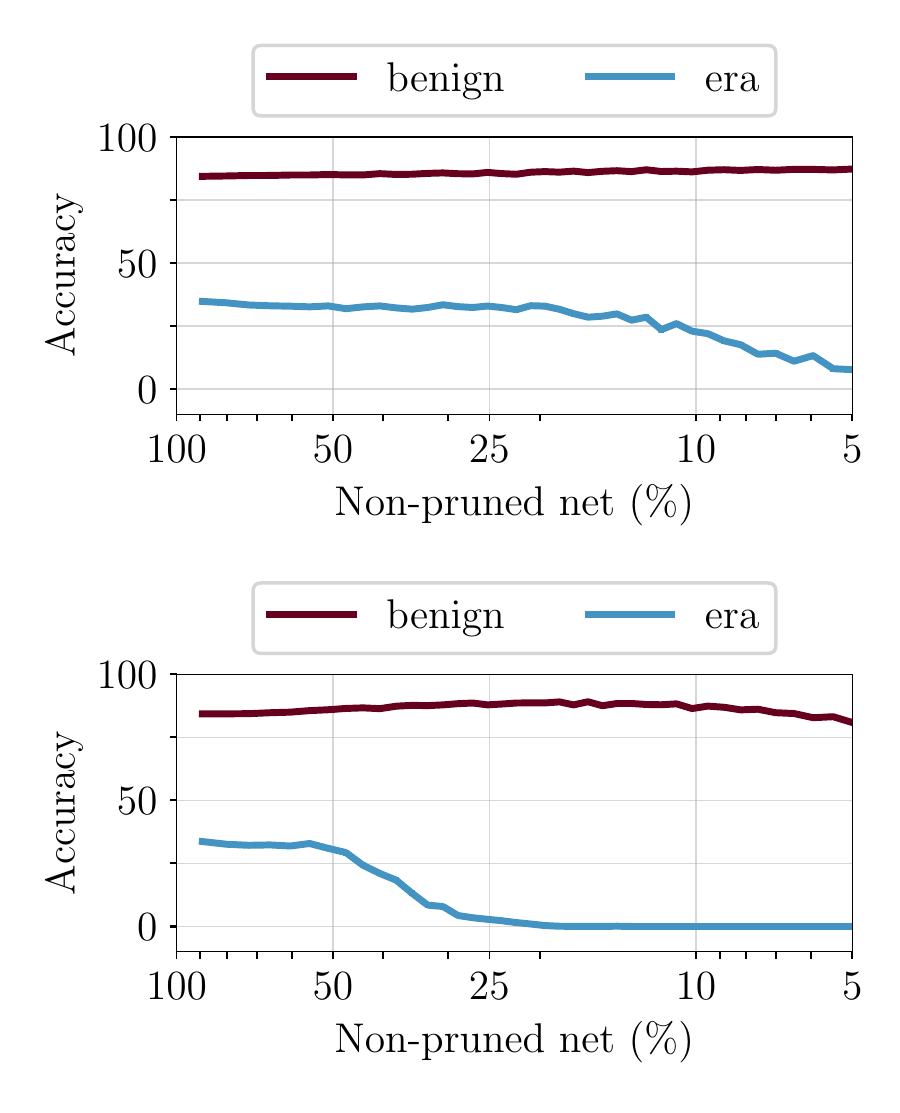}
        \caption{Adversarially trained model}
    \end{subfigure} 
    \caption{Natural training based fine-tuning for VGG16 network trained on CIFAR-10 dataset.}
    \label{fig:nat_finetune_vgg_full}
\end{figure*}

\pagebreak

\subsection{Fine-tuning based on Adversarial training.}\label{appsubsec:adv_finetune}
\begin{figure*}[!htb]
    \centering
    \begin{subfigure}[t]{0.48\textwidth}
        \centering
        \includegraphics[width=\linewidth]{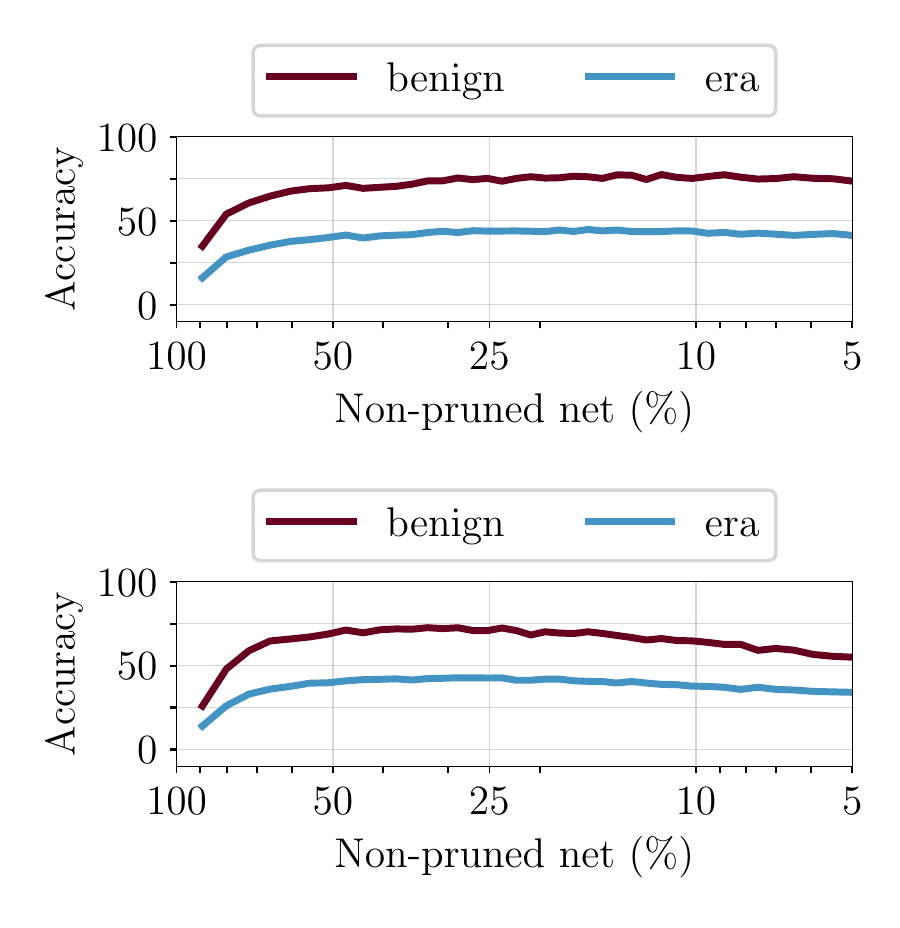}
        \caption{Naturally trained model}
    \end{subfigure}
    \begin{subfigure}[t]{0.48\textwidth}
        \centering
        \includegraphics[width=\linewidth]{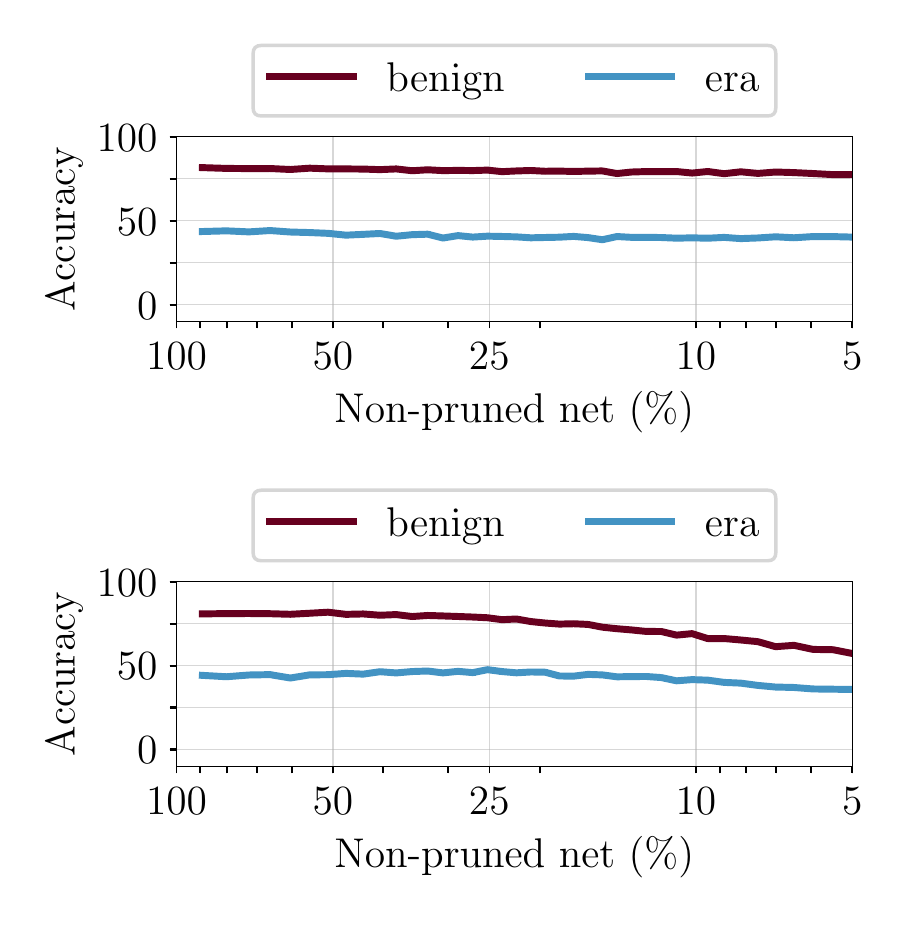}
        \caption{Adversarially trained model}
    \end{subfigure} 
    \caption{Adversarial training based fine-tuning for VGG16 network trained on CIFAR-10 dataset.}
    \label{fig:adv_finetune_vgg_full}
\end{figure*}

\section{Detailed results for provable robustness (\textit{vra}) relationship with network pruning.}
\begin{figure*}[!htb]
    \centering
    \begin{subfigure}[t]{0.75\textwidth}
        \centering
        \includegraphics[width=\linewidth]{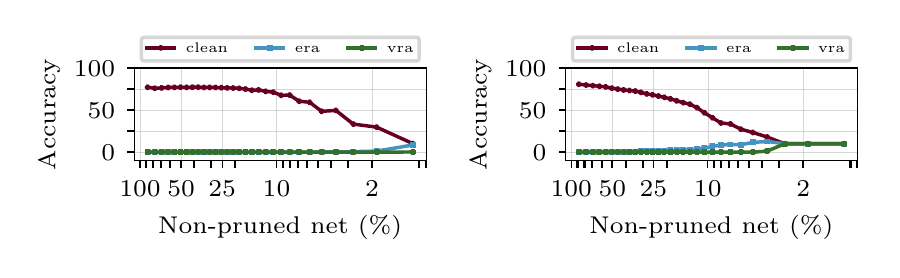}
        \caption{Naturally trained model, fine-tuned with MixTrain $k=1$ (original vra 0.0\%, acc 81.4\%, era 0.1\%, fine-tuning with k=1)}
        \label{fig:ma}
    \end{subfigure}
    
    \begin{subfigure}[t]{0.75\textwidth}
        \centering
        \includegraphics[width=\linewidth]{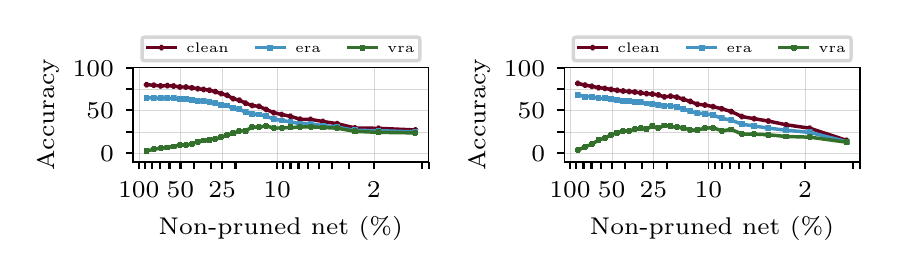}
        \caption{Adversarially trained model, fine-tuned with MixTrain $k=1$ (original vra 0.0\%, acc 83.2\%, era 69.3\%)}
        \label{fig:mb}
    \end{subfigure} 
    
    \begin{subfigure}[t]{0.75\textwidth}
        \centering
        \includegraphics[width=\linewidth]{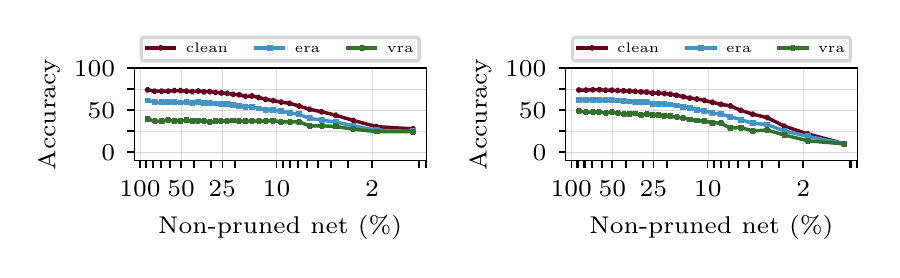}
        \caption{Robustly trained model, fine-tuned with MixTrain $k=1$ (original vra 51.0\%, acc 73.4\%, era 62.0\%, fine-tuning with k=1).}
        \label{fig:mc}
    \end{subfigure}
    
    \begin{subfigure}[t]{0.75\textwidth}
        \centering
        \includegraphics[width=\linewidth]{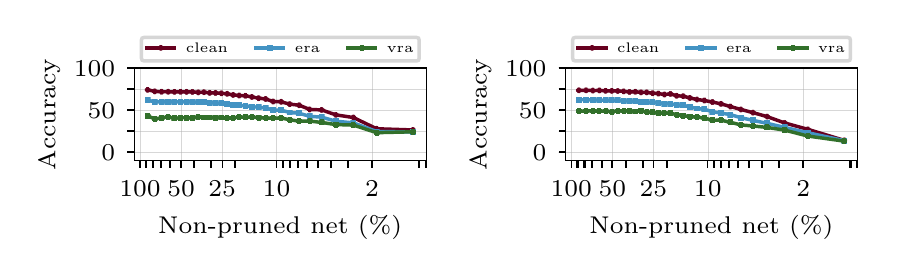}
        \caption{Robustly trained model, fine-tuned with MixTrain $k=1$ (original vra 51.0\%, acc 73.4\%, era 62.0\%, fine-tuning with k=10).}
        \label{fig:md}
    \end{subfigure}
    \caption{For \textit{CIFAR-Large} model, fine-tuning based on robust training using MixTrain with unstructured pruning (left) and structured pruning (right). The robust accuracy, both \textit{era} and \textit{vra} are reported with untargeted attacks.}
    \label{fig:cifar_large_mixtrain_full}
\end{figure*}

In Fig.~\ref{fig:cifar_large_mixtrain_full}, we show the detailed results for pruning with MixTrain $k=10$ on different types of pre-trained networks fine-tuned, including regularly trained network, adversarially robust trained network, and verifiably robust trained network. They share the same structure as \textit{CIFAR-Large}. We use MixTrain with $k=10$ to pre-train the verifiable robust network. For all of the pre-trained networks, their \textit{vra}, \textit{era}, and benign accuracy can be found in captions. 

\section{Training pruned models from scratch} \label{app:scratch_baseline}
In this section, we quantify the benefits of pruning by comparing it with pruned models which are trained from scratch (i.e., no pre-training and fine-tuning strategy is used). Table~\ref{table:scratch_clean} shows the advantages of latter approach when only no adversary is presented (i.e., only natural training is used). 

\begin{table}[!htb]
\caption{Comparing the benign accuracy of compact model resulted from the compression with compact model trained from scratch using natural training.}
\label{table:scratch_clean}
\centering
\begin{tabular}{ccccc}
\toprule
\multirow{2}{*}{Model} & \multirow{2}{*}{\begin{tabular}[c]{@{}c@{}}Pruning \\ ratio (\%)\end{tabular}} &  & \multicolumn{2}{c}{Benign accuracy (\%)} \\ \cline{4-5} 
 &  &  & scratch & fine-tuned  \\ \midrule
\multirow{3}{*}{VGG16} & 50 &  & 92.7 & \textbf{93.8} \\
 & 75 &  & 89.9 & \textbf{91.8}\\
 & 90 &  & 84.7 & \textbf{88.9}\\ \bottomrule
\end{tabular}
\end{table}


\end{document}